\documentclass[Journal,InsideFigs,letterpaper,NoLineNumbers]{ascelike-new}
\usepackage[utf8]{inputenc}
\usepackage[T1]{fontenc}
\usepackage{lmodern}
\usepackage{graphicx}
\usepackage[style=base,figurename=Fig.,labelfont=bf,labelsep=period]{caption}
\usepackage{amsmath}
\usepackage{newtxtext,newtxmath}
\usepackage[colorlinks=true,citecolor=black,linkcolor=black]{hyperref}
\usepackage{rotating}
\usepackage{graphicx}
\usepackage{array}
\usepackage{booktabs}
\usepackage{pdflscape}
\usepackage{tabularx} 
\usepackage{booktabs}
\usepackage{longtable}
\usepackage{subfigure}
\usepackage{adjustbox}
\usepackage{color,soul}
\usepackage{multirow}
\usepackage{tikz}
\usetikzlibrary{arrows.meta,positioning,fit,calc,shadows.blur}
\usepackage{rotating}

\usepackage{etoolbox}   
\robustify\cite         

\begin{document}

\title{\textsc{OpenConstruction}: A Systematic Synthesis of Open Visual Datasets for Data-Centric Artificial Intelligence in Construction Monitoring}

\author[1]{Ruoxin Xiong}
\author[2]{Yanyu Wang}
\author[3]{Jiannan Cai}
\author[4]{Kaijian Liu}
\author[5]{Yuansheng Zhu}
\author[6]{Pingbo Tang}
\author[7]{Nora El-Gohary}

\affil[1]{Assistant Professor, College of Architecture \& Environmental Design, Kent State University, Kent, OH, 44242, United States. Email: rxiong3@kent.edu}

\affil[2]{Assistant Professor, Bert S. Turner Department of Construction Management, Louisiana State University, Baton Rouge, LA, 70803, United States (corresponding author). Email: yanyuwang@lsu.edu}

\affil[3]{Assistant Professor, School of Civil \& Environmental Engineering, and Construction Management, The University of Texas at San Antonio, San Antonio, TX, 78249, United States. Email: jiannan.cai@utsa.edu}

\affil[4]{Assistant Professor, Department of Civil, Environmental, and Ocean Engineering, Stevens Institute of Technology, Hoboken, NJ, 07030, United States. Email: kliu24@stevens.edu}

\affil[5]{PhD Candidate, Department of Computing and Information Sciences, Rochester Institute of Technology, Rochester, NY, 14623, United States. Email: yz7008@g.rit.edu}

\affil[6]{Associate Professor, Department of Civil and Environmental Engineering, Carnegie Mellon University, Pittsburgh, PA, 15213, United States. Email: ptang@andrew.cmu.edu}

\affil[7]{Professor, Department of Civil and Environmental Engineering, University of Illinois at Urbana-Champaign, Urbana, IL, 61801, United States. Email: gohary@illinois.edu}

\maketitle

\begin{abstract}
The construction industry increasingly relies on visual data to support Artificial Intelligence (AI) and Machine Learning (ML) applications for site monitoring. High-quality, domain-specific datasets, comprising images, videos, and point clouds, capture site geometry and spatiotemporal dynamics, including the location and interaction of objects, workers, and materials. However, despite growing interest in leveraging these visual datasets, existing resources vary widely in size, data modalities, annotation quality, and representativeness of real-world construction conditions. A systematic review to categorize their data characteristics and application contexts is still lacking, limiting the community’s ability to fully understand the dataset landscape, identify critical gaps, and guide future directions toward more effective, reliable, and scalable AI applications in construction. To address this gap, this review conducts an extensive search of academic databases and open-data platforms, yielding 51 publicly available visual datasets that span the 2005–2024 period. These datasets are categorized using a structured data schema covering (i) data fundamentals (e.g., size and license), (ii) data modalities (e.g., RGB, point cloud, thermal), (iii) annotation frameworks (e.g., bounding boxes, keypoints), and (iv) downstream application domains (e.g., safety, progress tracking). This study synthesizes these findings into an open-source catalog, \textsc{OpenConstruction}, supporting data-driven method development. Furthermore, the study discusses several critical limitations in the existing construction dataset landscape and presents a roadmap for future data infrastructure anchored in the Findability, Accessibility, Interoperability, and Reusability (FAIR) and domain-specific principles. By reviewing the current landscape and outlining strategic priorities, this study supports the advancement of data-centric solutions in the construction sector.
\end{abstract}

\section{Introduction}
\label{sec:intro}
Construction projects are highly dynamic and complex, with site conditions that change rapidly and pose ongoing challenges for safety, productivity, and quality \cite{abioye2021artificial}. Construction monitoring has therefore become a central function in construction practice, involving systematic collection and analysis of site data to assess quality, track progress, examine safety, manage resources, and verify compliance \cite{kim2020visual}. Advances in computer vision and foundation models have expanded the role of artificial intelligence (AI) in this context, enabling accurate and scalable automated interpretation of visual data and supporting applications in progress monitoring, safety management, and equipment and material tracking \cite{paneru2021computer,li2023towards}. These tasks and applications are powered by diverse streams of visual data, including RGB images, videos, RGB-D streams, thermal imagery, and 3D point clouds, that capture site conditions and allow algorithms to interpret context, track progress, and detect risks \cite{mostafa2021review}.

Transforming these raw data streams into reliable AI applications requires well-curated and domain-specific datasets, as they provide training and evaluation materials \cite{rabbi2024ai}. Open-access datasets play a critical role in this process by offering shared benchmarks, allowing reproducibility, and fostering cumulative progress through community-driven innovation \cite{bilal2016big}. However, in construction, the availability of such datasets remains limited due to governance restrictions, proprietary and privacy concerns, and the cost and effort required for release \cite{wang2023characterizing}. These barriers restrict the scale, diversity, and accessibility of available data, slowing the advancement of robust and generalizable AI solutions in construction monitoring.

Even when datasets are publicly released, their use in construction monitoring is complicated by fragmentation and lack of systematic organization \cite{regona2022opportunities,wu2021towards}. Existing resources are dispersed across sites and repositories and are generally developed for specific monitoring tasks and applications, such as activity recognition, safety compliance, equipment tracking, or progress estimation \cite{wang2021characterizing,martinez2019scientometric}. They differ in modalities (e.g., RGB, depth, thermal, or 3D point cloud), spatial and temporal resolution, and annotation methods (e.g., bounding boxes and segmentation masks). Furthermore, essential descriptors, such as dataset size, capture conditions, project phase, sensor configurations, annotation schema, and licensing, are reported inconsistently and often scattered across publications and repositories \cite{bilal2016big}. Without a consistent and systematic cataloging framework, researchers face difficulties in discovering and reusing relevant datasets, comparing results between studies, and identifying systematic gaps in coverage \cite{wang2023characterizing,wilkinson2016fair}.

This paper addresses these gaps by providing systematic review and synthesis of open-access visual datasets for construction monitoring. In line with scientific data sharing practices \cite{wilkinson2016fair}, we examine datasets that are publicly reachable and available for research use across a range of licensing conditions, and we offer clear documentation to support responsible and compliant reuse. Specifically, our contributions are threefold. First, we systematically identify and catalog publicly available datasets through a comprehensive review of scholarly and open-data platforms. Second, we analyze dataset fundamentals, modalities, annotation frameworks, and application domains to provide a structured perspective on their characteristics and limitations. To support community-driven development, we release an open-source catalog, \textsc{OpenConstruction} ({\href{https://github.com/ruoxinx/OpenConstruction-Datasets}{https://github.com/ruoxinx/OpenConstruction-Datasets}}), which indexes these resources for transparent access. Finally, we synthesize the critical gaps of current datasets and propose a roadmap for future data initiatives, focusing on multimodal sensing, ontology standardization, benchmark protocol, data governance, and community engagement. Through this synthesis, this review aims to strengthen the data foundations required for robust, trustworthy, and scalable AI solutions in construction monitoring.

\section{Background}
\label{sec:review}
This section reviews advances in vision-based construction monitoring, outlines data requirements and development strategies to support these AI applications, and examines the state of data infrastructure, governance, and sharing barriers that shape their availability and reuse.

\subsection{Advances in Vision-Based Construction Monitoring}
Vision-based construction monitoring refers to the use of image, video, and 3D sensing technologies to capture site conditions and enable automated analysis that supports safety, productivity, quality, and resource management \cite{kim2020visual}. Recent studies have demonstrated a variety of applications. Progress monitoring has been advanced through image-based documentation, drone surveys, and Scan-to-Building Information Modeling (BIM) workflows, where as-built point clouds are compared against design models to verify construction accuracy and progress \cite{ekanayake2021computer}. Safety monitoring has applied computer vision to detect hazards, recognize unsafe worker behavior, and verify compliance \cite{seo2015computer}. Visual data also support equipment and material tracking, enabling real-time analysis of resource location and utilization, while quality assurance and control rely on vision-based inspections to identify defects and deviations \cite{sherafat2020automated}.

These applications have been made possible by advances in AI architectures. Convolutional neural networks (CNNs), transformers, and vision–language models achieve state-of-the-art performance in detection, segmentation, tracking, and captioning, outperforming traditional feature-engineering approaches \cite{pan2021roles}. Such capabilities demonstrate the potential of AI to deliver scalable, fine-grained, and real-time insights that benefit construction management.

\subsection{Data Requirements and Dataset Development for Construction Monitoring}
The reliability of AI models is increasingly recognized to be constrained not only by architecture alone, but also by the availability and quality of the underlying data \cite{whang2023data}. This reflects the principles of data-centric AI, which prioritize dataset quality, diversity, and standardization as foundations for robust and generalizable systems \cite{zha2023data}. For construction monitoring, where variability in project type, scale, and environment is high, well-curated datasets are particularly essential \cite{xu2021computer}.

Developing such datasets involves three key processes. First, \textit{data collection} typically employs fixed cameras, drones, LiDAR, or mobile devices, but variability in layout, weather, and lighting often hinders consistency \cite{bilal2016big}. Synthetic data generation through BIM, simulation platforms, and generative models provides a complementary strategy to represent rare or hazardous scenarios at scale \cite{hong2021synthetic,lee2023game,jiang2023building,kim2024image}. Second, \textit{data labeling} is required to produce training examples, but manual annotation of workers, equipment, and hazards is resource-intensive. Semi-supervised, weakly supervised, and active learning methods have been employed to reduce redundant effort and focus human expertise on critical samples \cite{wang2019crowdsourced}. Third, \textit{data preparation}, including normalization, metadata documentation, augmentation, and partitioning into training, validation, and test sets, ensures fairness, comparability, and reproducibility in model evaluation \cite{pal2021deep}.

\subsection{Data Infrastructure and Sharing Barriers in Construction Monitoring}
Despite progress in dataset development, the construction sector has not yet fully established infrastructures to support open-access data availability and reuse \cite{wang2023characterizing}. Proprietary ownership of project records and contractual arrangements can limit opportunities for sharing, while considerations of worker privacy may restrict the release of imagery containing identifiable individuals \cite{regona2022opportunities}. Even when data sharing is intended, the efforts required for curation, annotation, anonymization, and long-term hosting present practical challenges, particularly for academic and non-profit initiatives \cite{wu2021towards}.

A related issue is the limited presence of centralized repositories and standardized governance frameworks \cite{karji2022identification}. In domains such as medical imaging and autonomous driving, community-wide repositories and policies have established standards for the indexing, documentation, and long-term accessibility of data sets \cite{diaz2021data,liu2024survey}. In comparison, construction has fewer comparable infrastructures, and important metadata, such as capture conditions, project phase, sensor configuration, and licensing terms, are sometimes reported inconsistently or dispersed between publications and repositories \cite{yang2022datasets}. This variability may complicate dataset discovery, reduce transparency, and restrict reproducibility and reuse.

Together, these factors suggest the value of more systematic efforts to organize existing resources. By consolidating dispersed resources, characterizing their attributes in a structured manner, and identifying coverage gaps, such efforts can provide the transparency, comparability, and guidance needed to advance data-centric AI applications in construction monitoring.

\section{Review of Open Visual Datasets in Construction Monitoring}
\label{sec:method}
\begin{figure}[ht]
    \centering
    \includegraphics[width=0.75\linewidth]{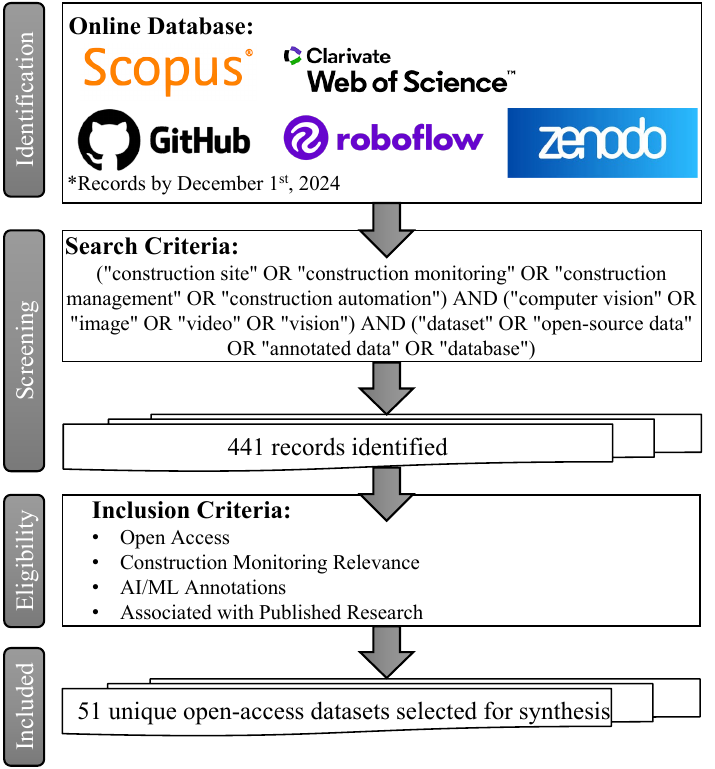}
    \caption{Framework for the search, screening, and synthesis of open-access visual datasets in construction monitoring}
    \label{fig:frame}
\end{figure}

To ensure a comprehensive review of relevant datasets, this study used a structured multistep search strategy (Fig.~\ref{fig:frame}). The process began with the formulation of a targeted Boolean query designed to systematically identify open-access visual datasets in construction monitoring.

The search string consists of three primary categories: (1) \textit{domain context terms:} to focus on the construction industry and related activities, including terms such as \texttt{"construction site"}, \texttt{"construction monitoring"}, \texttt{"construction management"}, and \texttt{"construction automation"}, (2) \textit{visual data and computer vision terms:} to identify datasets containing visual content suitable for AI/ML workflows, including \texttt{"computer vision"}, \texttt{"image"}, \texttt{"video"}, and \texttt{"vision"}, and (3) \textit{data accessibility and format:} to target records referring to datasets and their availability in open-access formats, including \texttt{"dataset"}, \texttt{"open-source data"}, \texttt{"annotated data"}, and \texttt{"database"}. The final Boolean query employed was the following:
\begin{quote}
(\texttt{"construction site” OR "construction monitoring” OR "construction management” OR "construction automation”}) AND (\texttt{"computer vision” OR "image” OR "video” OR "vision”}) \ AND (\texttt{"dataset” OR "open-source data” OR "annotated data” OR "database”}) 
\end{quote}

Searches were conducted on academic databases (Scopus~\cite{Scopus} and Web of Science~\cite{WebOfScience}) and public repositories (GitHub~\cite{GitHub}, Zenodo~\cite{Zenodo}, and Roboflow~\cite{roboflow2024}). To prevent duplication across sources, the metadata for all identified datasets, including dataset name, release date, data type, collection method, categories, size, resolution, annotation types, license, applications, and associated publications, was systematically cross-checked to identify repeated entries. This process yielded 441 candidate records. Datasets were then screened according to the following inclusion criteria:
\begin{enumerate}
    \item \textbf{Open Accessibility:} Publicly available for academic and research purposes through direct download, repository access, or standard registration procedures. We include datasets across the spectrum of licensing conditions, from permissive (e.g., CC0, CC-BY) to academic-restricted (e.g., CC-BY-NC). Datasets with proprietary restrictions, restrictive licensing barriers, or extended negotiation processes were excluded.
    \item \textbf{Relevance to Construction Monitoring:} Containing visual content representing construction environments, including sites, materials, equipment, workers, and associated activities.
    \item \textbf{AI/ML-Ready Annotations:} Providing machine-readable annotations suitable for computer vision tasks, such as bounding boxes, segmentation masks or textual descriptions.
    \item \textbf{Dataset Documentation:} Supported by a peer-reviewed publication or technical report describing dataset creation, structure, and use. In cases lacking direct documentation, datasets were considered only if their quality and utility had been validated in peer-reviewed research. To ensure accessibility, only sources published in English were included.
\end{enumerate}

Applying these criteria resulted in a final selection of 51 unique open-access datasets. These span a wide temporal range (2005–2024), cover diverse geographic regions and sensing modalities, and address various tasks relevant to construction monitoring. The dataset search and identification process was finalized as of December 1, 2024, establishing the temporal scope of this review.

\section{Results of the Dataset Review}
\label{sec:results}
This section characterizes the identified datasets, focusing on their temporal evolution, geographic distribution, modalities, annotated objects, supported tasks, and domain applications.

\subsection{Dataset Overview}
Through the systematic search and screening process, 51 open-access visual datasets relevant to construction monitoring were identified that met the inclusion criteria. Information was extracted from dataset documentation, associated publications, and metadata. A detailed list of all datasets, covering publication year, data type, collection methods, size, number of classes, resolution, annotation types, construction elements, licensing, and access mechanisms, is provided in Appendix~\ref{app:datasets}. This consolidated resource facilitates informed dataset selection for researchers and practitioners.

\begin{figure}[htbp]
\centering
\subfigure[Temporal distribution (2015–2024) \label{fig:temporal}]
{\includegraphics[width=0.49\linewidth]{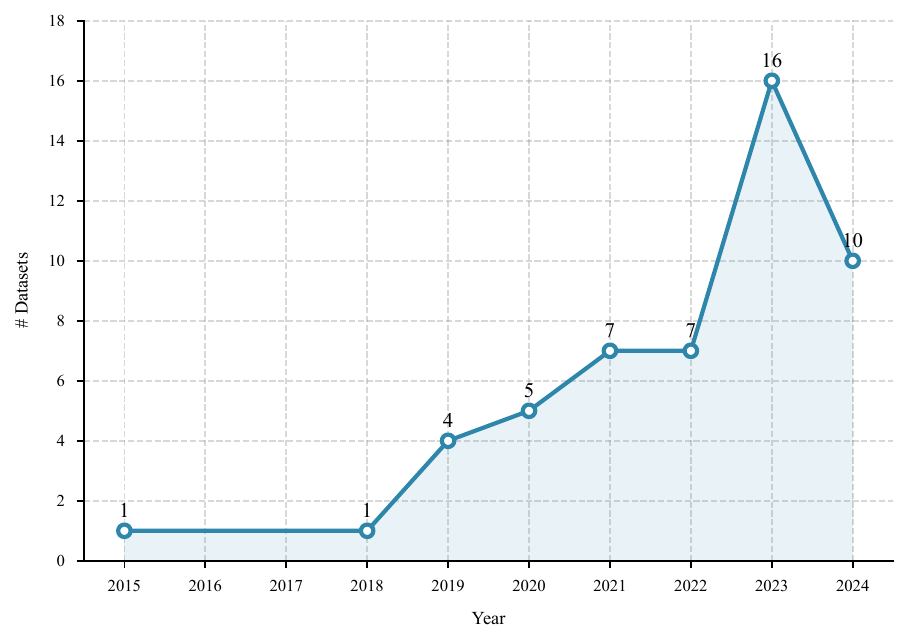}}
\subfigure[Geographic distribution of data collection locations \label{fig:geo}]
{\includegraphics[width=0.49\linewidth]{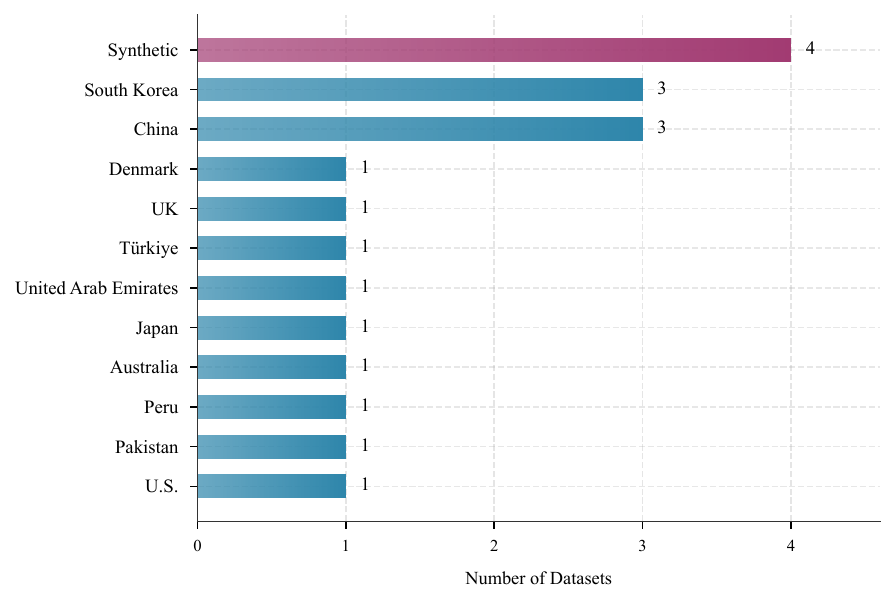}}
\caption{Temporal and geographic distribution of the identified construction visual datasets. Datasets without specified collection locations are excluded.}
\label{fig:tools}
\end{figure}

The temporal distribution of the datasets spans 2015–2024, with a marked increase in availability during recent years (Fig.~\ref{fig:temporal}). Prior to 2020, fewer than five datasets were released annually, whereas by 2023 the number exceeded ten per year. This growth reflects the increasing integration of data-centric AI into the construction industry, the expansion of open platforms and the proliferation of low-cost imaging technologies, including unmanned aerial vehicles (UAVs) and smartphones. The increasing availability of datasets has supported applications in safety monitoring, progress tracking, and predictive analytics within construction management \cite{abioye2021artificial}.

Geographic analysis (Fig.~\ref{fig:geo}) indicates a wide international participation, with data collection efforts spanning Europe, North America, East Asia, South America, the Middle East, and Oceania. This distribution highlights both the global relevance of AI/ML techniques for construction monitoring and the collaborative efforts of research communities to advance this domain.

\subsection{Dataset Characterization}
\label{sec:charac}
\begin{figure}[ht]
    \centering
    \includegraphics[width=0.9\linewidth]{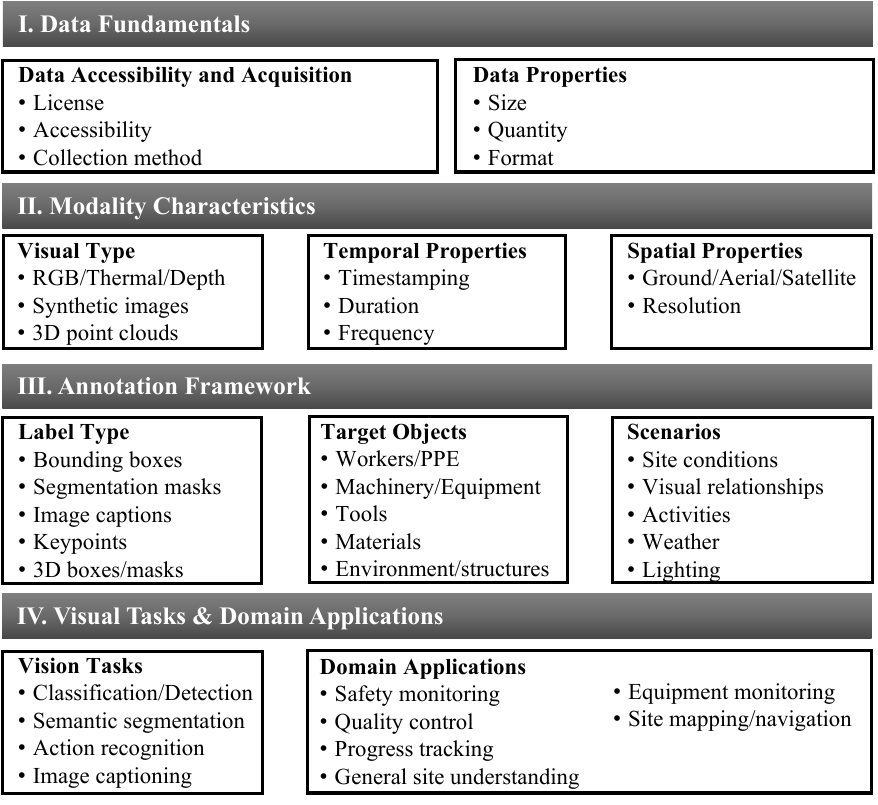}
    \caption{Schema for characterizing open visual datasets in construction monitoring}
    \label{fig:taxonomy}
\end{figure}

The selected datasets are systematically organized using a data schema (Fig.~\ref{fig:taxonomy}), which builds on established frameworks in dataset characterization~\cite{yang2022datasets,nie2022open}. The schema distinguishes four complementary categories: (1) \textit{Data Fundamentals} capture core attributes that define each dataset, including storage size, number of instances, and file format, along with acquisition and accessibility information such as collection method, licensing, and access conditions. (2) \textit{Modality Characteristics} describe the types of visual data (e.g., RGB imagery, 3D point clouds, RGB-D, thermal, synthetic), temporal properties (timestamping, duration, frequency), and spatial properties (aerial or ground perspectives, resolution). (3) \textit{Annotation Framework} specify the annotation types provided (e.g., bounding boxes, segmentation masks, captions, keypoints), the annotated entities (e.g., workers, personal protective equipment (PPE)) and the contextual scenarios (site conditions, activity types). (4) \textit{Visual Tasks \& Domain Applications} link dataset attributes to specific AI-driven tasks such as classification, detection, segmentation, and captioning, as well as their domain applications such as safety monitoring, quality control, and progress monitoring.

\subsubsection{Dataset Size and Accessibility in Datasets}
\begin{figure}[htbp]
    \centering
    \subfigure[Distribution of dataset sizes\label{fig:size_count}]{
        \includegraphics[width=0.48\linewidth]{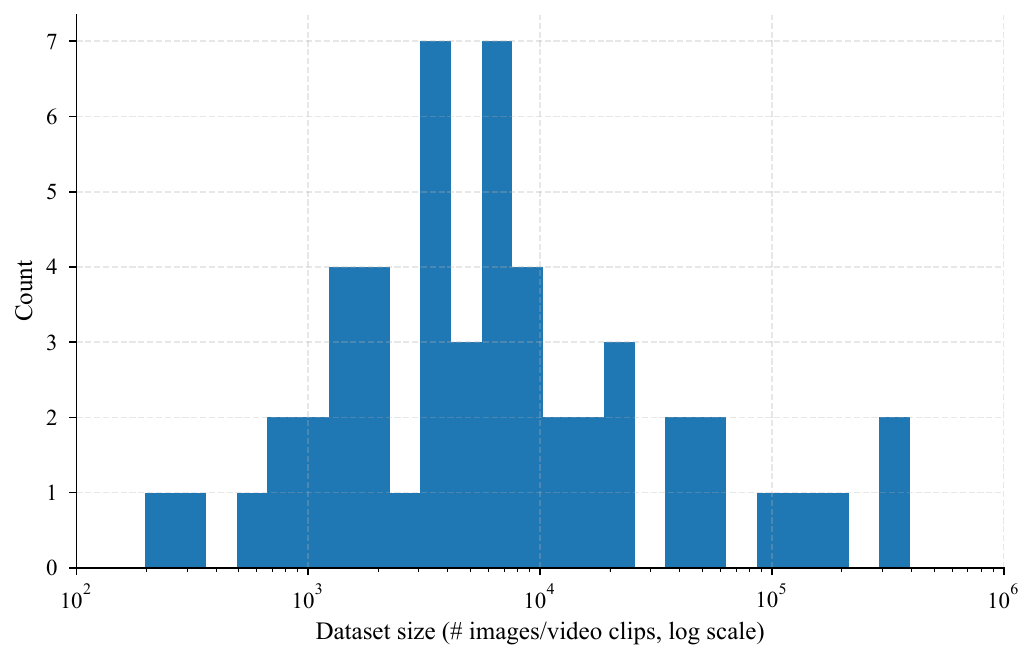}
    }
    \subfigure[Dataset sizes by task type (bubble size = classes)\label{fig:size_task}]{
        \includegraphics[width=0.48\linewidth]{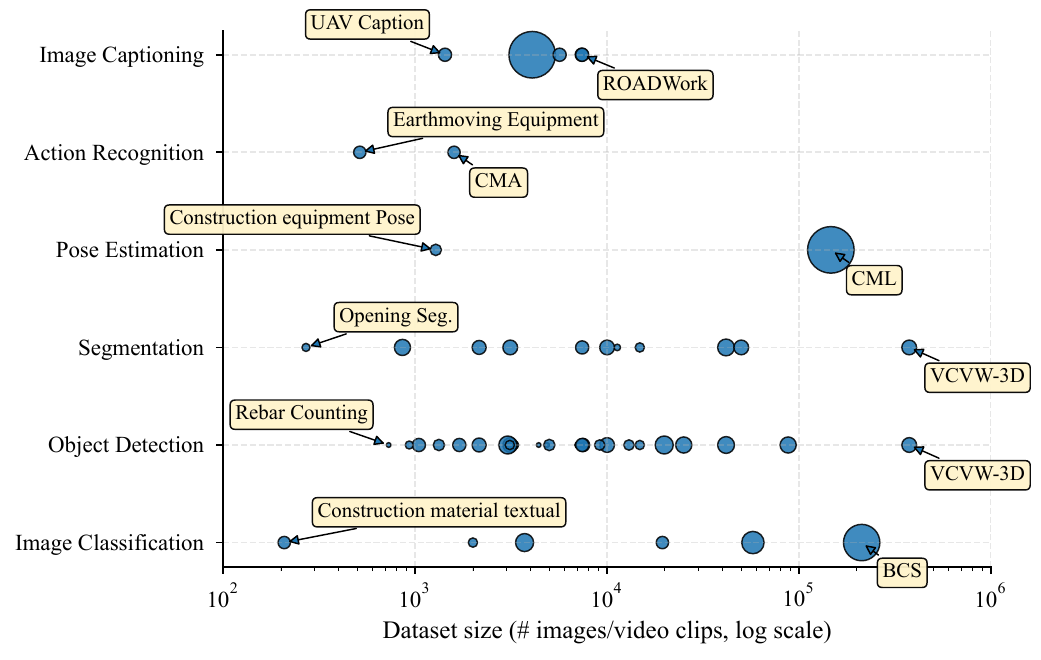}
    }
    \caption{Distribution of dataset sizes and their coverage across tasks}
    \label{fig:size}
\end{figure}

\paragraph{Dataset Size} The identified datasets vary significantly in size, ranging from a few hundred images to over 300,000 samples. The overall distribution (Fig.~\ref{fig:size_count}) reveals that the majority (56\%) contains between $10^{3}$ and $10^{4}$ images or video clips, representing the dominant range of dataset availability. Smaller datasets with fewer than $10^{3}$ samples account for 13\%, typically targeting narrowly defined tasks such as rebar inspection or building element segmentation. Only 7\% exceed $10^{5}$ instances (e.g., BCS~\cite{cheng2024large}, VCVW-3D~\cite{ding2024virtual}, CML~\cite{tian2022construction}).

Task-level analysis (Fig.~\ref{fig:size_task}) shows differences in dataset scale between tasks. Image classification datasets are generally larger, often exceeding $10^{4}$ samples, consistent with their need for diverse coverage of materials and building categories. Object detection and segmentation datasets span the widest range, from a few hundred to more than $10^{5}$ images, reflecting trade-offs between annotation complexity and dataset scale. Higher-level tasks such as pose estimation, action recognition, and image captioning are typically smaller (below $10^{4}$), given the resource-intensive nature of annotation. Two datasets, CML~\cite{tian2022construction} and Earthmoving Equip. Tracking~\cite{roberts2019end}, are distinctive in providing video clips rather than static imagery, thus contributing to the temporal dynamics essential for activity recognition.

\begin{figure}[htbp]
\centering
\subfigure[Distribution of license types \label{fig:license}]
{\includegraphics[width=0.49\linewidth]{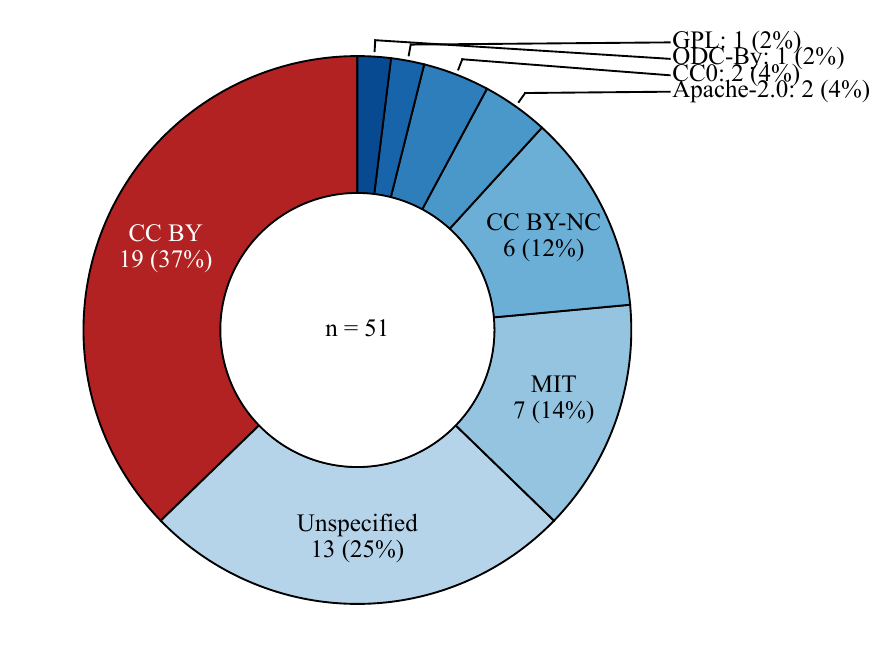}}
\subfigure[Distribution of data modalities \label{fig:license_modality}]
{\includegraphics[width=0.49\linewidth]{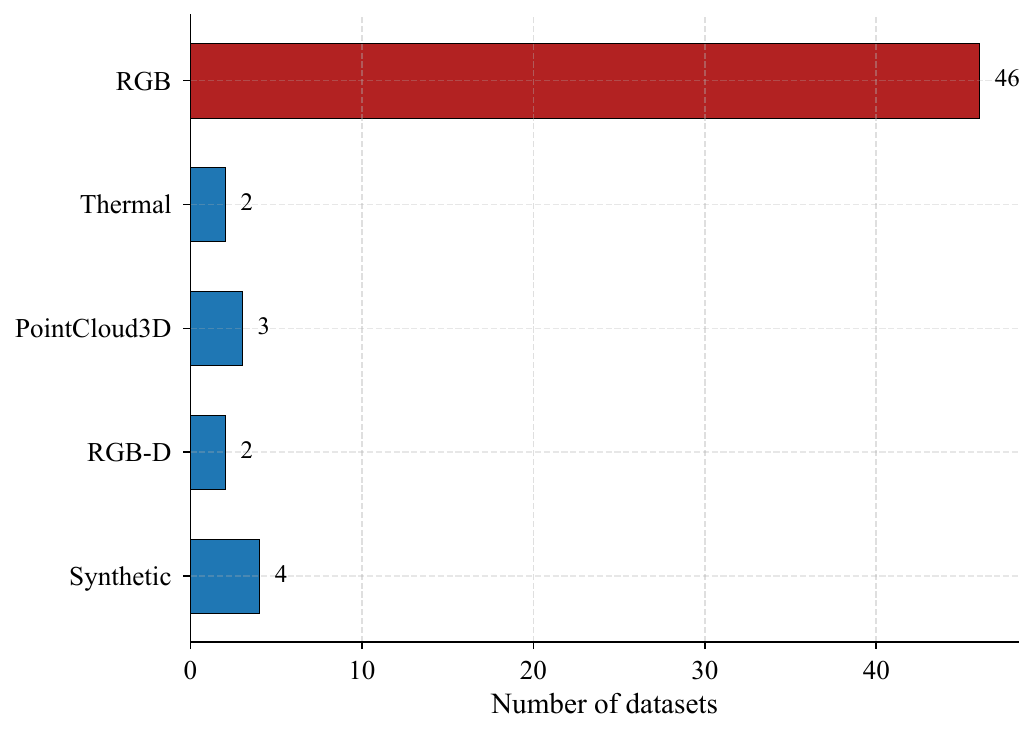} }
\caption{Overview of dataset characteristics in license types and data modalities (note: some datasets include multiple modalities).}
\end{figure}

\paragraph{Dataset Accessibility}
Licensing terms play a critical role in determining the reusability, interoperability, and downstream integration of visual datasets in applications. Across the 51 identified datasets, the licensing landscape is highly heterogeneous (Fig.~\ref{fig:license}). The most common permissive license is CC BY (19 datasets, 37\%), followed by MIT (7 datasets, 14\%) and CC0 (2 datasets, 4\%), all of which support broad redistribution and integration into downstream workflows. Examples include the \textit{MINC-2500} dataset \cite{bell2015material} (CC BY 4.0) and the \textit{Hardhat-Wearing Dataset} \cite{wang2020hardhat} (CC0 1.0), which provide accessible resources for benchmarking and model development. More restrictive terms appear less frequently, including CC BY-NC (6 datasets, 12\%), Apache-2.0 (2 datasets, 4\%), GPL (1 datasets, 2\%), and ODC-BY (1 dataset, 2\%). A substantial portion, 13 datasets (25\%), offers publicly accessible links but lacks a standard or formal license. Overall, the diversity of licensing underscores the need for clearer guidance and more standardized release practices in the construction monitoring research community.

\subsubsection{Data Modalities and Acquisition Methods in Datasets}
\begin{figure}[htbp] 
\centering 
\includegraphics[width=0.98\textwidth]{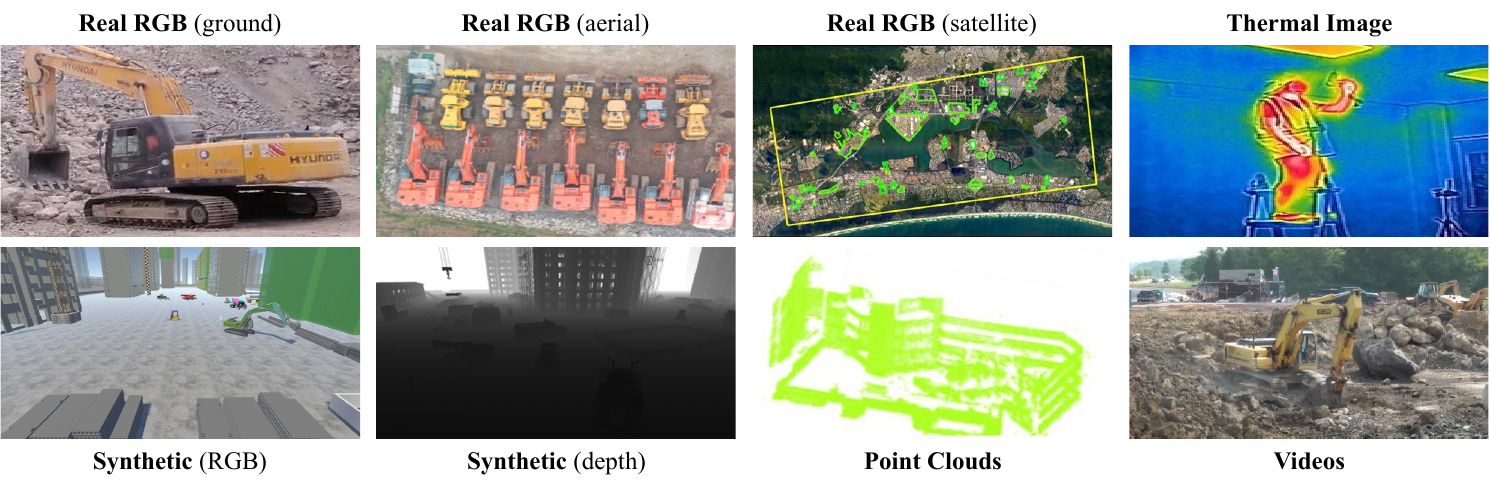} 
\caption{Examples of dataset modalities, including real RGB, thermal imagery, synthetic data, point clouds, and video clips. Sources (left to right, top to bottom): \textit{Equip. Pose} \cite{luo2020full}; \textit{AIDCON} \cite{ersoz2024aidcon}; \textit{SMART Dataset} \cite{goldberg2023automated}; \textit{Thermal Safety Dataset} \cite{yirong2023thermal}; \textit{VCVW-3D} \cite{ding2024virtual}; \textit{VCVW-3D} \cite{ding2024virtual}; \textit{PC-Urban} \cite{ibrahim2021annotation}; \textit{Earthmoving Equip. Tracking} \cite{roberts2019end}.} 
\label{fig:data_types} 
\end{figure}

\paragraph{Data Modalities}
The identified datasets employ a variety of sensing modalities that reflect different acquisition technologies and application requirements. Representative examples of primary modalities are illustrated in Fig.~\ref{fig:data_types}, and their distribution across the dataset collection is summarized in Fig.~\ref{fig:license_modality}. RGB imagery is by far the most common modality, appearing in 46 of the 51 datasets. These datasets cover multiple perspectives: ground-level images (e.g., \textit{Construction Equipment Pose} \cite{luo2020full}), aerial views captured by UAVs (e.g., \textit{AIDCON} \cite{ersoz2024aidcon}), and satellite imagery (e.g., \textit{SMART dataset} \cite{goldberg2023automated}). RGB serves as the foundation for most object detection, segmentation, and classification tasks.

Other modalities are less common but provide valuable complementary perspectives. Synthetic datasets (8\%), generated using simulation platforms or game engines, include examples such as \textit{VCVW-3D} \cite{ding2024virtual}, which produces RGB and depth data for controlled experimentation and rare scenario modeling. 3D point clouds (6\%) obtained through LiDAR or multi-sensor systems support geometric reasoning and site mapping, as in the \textit{PC-Urban} dataset \cite{ibrahim2021annotation}. RGB-D datasets (4\%) integrate color and depth information for tasks such as 3D pose estimation and activity recognition. Thermal imagery (4\%) supports safety monitoring and worker activity analysis under low-light or occluded conditions. Finally, a small set of datasets consists of video clips, such as the \textit{Earthmoving Equipment Tracking Dataset} \cite{roberts2019end}, which supports action recognition and temporal activity analysis.

\paragraph{Data Acquisition}
Visual data for construction monitoring are collected through various acquisition methods. Ground-level RGB imagery, obtained via handheld or stationary cameras, provides detailed viewpoints with low equipment costs and ease of deployment. However, such data often suffers from limited coverage, occlusions, and high labor requirements for consistent capture \cite{mostafa2021review}. UAV-based aerial imagery expands spatial reach and enables rapid overhead acquisition with relatively affordable hardware, but requires skilled operators and is subject to weather conditions and flight regulations \cite{zhou2018unmanned}. Satellite imagery supports regional to global monitoring and offers access to historical archives, but its lower spatial resolution, limited temporal frequency, and high licensing costs reduce its practicality for project-level monitoring \cite{goldberg2023automated}. 

Synthetic data, generated through simulation or game engine environments, allow for scalable and customizable dataset creation at low cost, particularly useful for controlled experimentation. However, such data often lacks the richness of real-world variability and risk introducing domain bias when applied to operational environments \cite{hong2021synthetic}. Thermal imagery, captured using infrared cameras, supports safety analysis and anomaly detection tasks. Although increasingly affordable, thermal sensors remain more costly than RGB cameras, require expert interpretation, and are sensitive to ambient conditions \cite{wu2023thermal}. Video recordings, acquired with handheld, fixed, or UAV-mounted cameras, capture temporal dynamics and process flows, providing richer context than still images. However, these advantages come with higher storage demands, greater analytic complexity, and increased privacy concerns \cite{gong2010computer}. Point clouds generated by LiDAR or photogrammetry offer high-fidelity 3D representations. However, their deployment requires expensive sensors, substantial storage, and intensive processing demands, which has limited their representation in these open-access datasets \cite{wang2019applications}.

\subsubsection{Annotations, Target Objects and Construction Activities in Datasets}
\paragraph{Annotation Types}
The reviewed datasets adopt diverse annotations that vary in granularity and application. Bounding boxes are the most common and are used for object detection by enclosing objects within predefined boxes. For example, the \textit{SCUT-HEAD} dataset \cite{peng2018detecting} employs bounding boxes to detect workers’ heads, enabling safety monitoring for helmet use. More detailed annotation is provided in the pixel-level segmentation, used in 22\% of the datasets. This method assigns a class to each pixel, supporting fine-grained tasks such as defect detection and material assessment. The \textit{Concrete Crack Dataset} \cite{liu2019deep}, for example, applies segmentation to identify and analyze structural cracks for inspection and maintenance.

Two datasets (4\%) include 2D or 3D keypoints, marking distinctive objects or human joints to capture pose and movement. Such annotations support activity recognition and ergonomic analysis. The \textit{Construction Equipment Pose Dataset} \cite{luo2020full} uses keypoints to characterize the orientation of excavators, contributing to research on autonomous construction robotics. A smaller set of datasets (4\%) incorporate 3D annotations, which provide explicit geometric information for spatial reasoning tasks such as BIM reconstruction and navigation. The \textit{VCVW-3D Dataset} \cite{ding2024virtual} exemplifies this category, offering extensive 3D labels for site objects and layouts.

\paragraph{Target Objects and Activities}
Target objects in the datasets span workers, equipment, tools, materials, and environmental characteristics (Table~\ref{tab:target_objects}). PPE and worker-related classes account for 35\% of datasets, focusing on helmets, vests, harnesses, and glasses to support safety compliance and injury prevention \cite{wu2019automatic,wang2020hardhat}. Machinery appears in 20\% of datasets, with categories such as excavators, cranes, and trucks supporting utilization monitoring, access control, and workflow analysis \cite{xiao2021development}. The \textit{Earthmoving Equipment Tracking Dataset} \cite{roberts2019end}, for example, records detailed machinery operations for planning and oversight. Material classes (e.g., bricks, concrete, steel) support quality control and defect detection \cite{mengiste2024transfer}, while environmental features (e.g., cracks and dust) facilitate site-condition and structural-health assessment \cite{lu2022computer}.

\begin{table}[htbp]
\centering
\caption{Summary of target objects in identified datasets}
\label{tab:target_objects}
\begin{adjustbox}{max width=\textwidth}
\begin{tabular}{@{}p{4cm}p{12cm}@{}}
    \toprule
    \textbf{Category} & \textbf{Target Objects} \\
    \midrule
    Worker/PPE & 
    Person (or worker), Head, Hardhat (or helmet), Colored hardhat (yellow, white, red, blue), Safety harness, Vest, Safety glasses \\ \midrule
    Machinery/Equipment & 
    Excavator, Compactor, Dozer, Grader, Dump truck, Concrete mixer truck, Wheel loader, Backhoe loader, Tower crane, Mobile crane, Vehicle crane, Hanging hook, Roller, Truck, Loader, Pump truck, Pile driver, Skid steer, Car, Particle monitors, Spraying car, Cutter, Electric box, Hopper, Spraying sink \\ \midrule
    Tools & 
    Bucket, Cord reel, Hammer, Tacker, Scaffold, Handcart \\ \midrule
    Materials & 
    Brick, Carpet, Ceramic, Concrete, Fabric, Foliage, Glass, Gypsum, Metal, Mirror, Paper, Plastic, Plasterboards, Polished stone, Stone, Tile, Wallpaper, Water, Wood, Pipes, Rebar, Circular pipes, Square tubes, I-beams, Precast components, Concrete Masonry Units (CMU) wall, Chiseled concrete, Concrete, Gypsum, Mesh, Coat plaster \\ \midrule
    Environment/Structures & 
    Dust, Building, Slogan, Fence, Concrete crack \\ 
    \bottomrule
\end{tabular}
\end{adjustbox}
\end{table}

In addition to static objects, around 20\% of datasets provide activity annotations, ranging from atomic actions to complex multi-step processes. These enable models for productivity measurement, workflow monitoring, and automated documentation \cite{sherafat2020automated}. Basic actions such as standing, walking, or ladder climbing, as in the \textit{CMA Dataset} \cite{yang2023transformer}, are applied in safety and productivity studies. Higher-level construction processes, such as masonry or rebar work captured in the \textit{Construction-Activity-Scenes Image Captioning Dataset} \cite{liu2020manifesting}, support detailed workflow and progress analysis.

\subsubsection{Vision Tasks and Domain Applications in Datasets}
\begin{figure}[ht] 
\centering 
\includegraphics[width=0.98\textwidth]{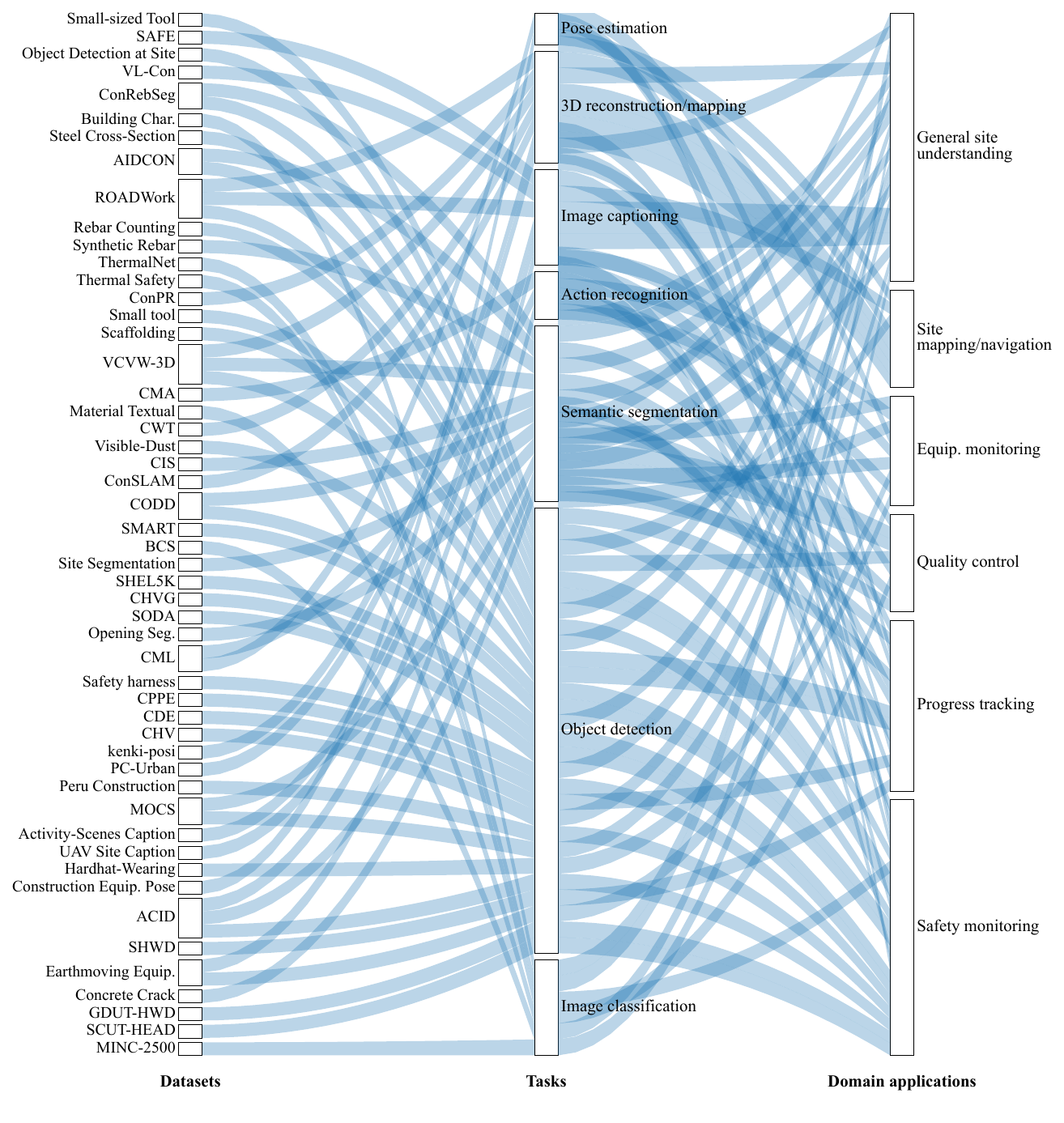} 
\caption{Mapping of datasets to supported computer vision tasks and domain applications in construction monitoring. Dataset abbreviations correspond to Appendix~\ref{app:datasets}.}
\label{fig:data_task_app} 
\end{figure}

\begin{figure}[ht] 
\centering 
\includegraphics[width=0.98\textwidth]{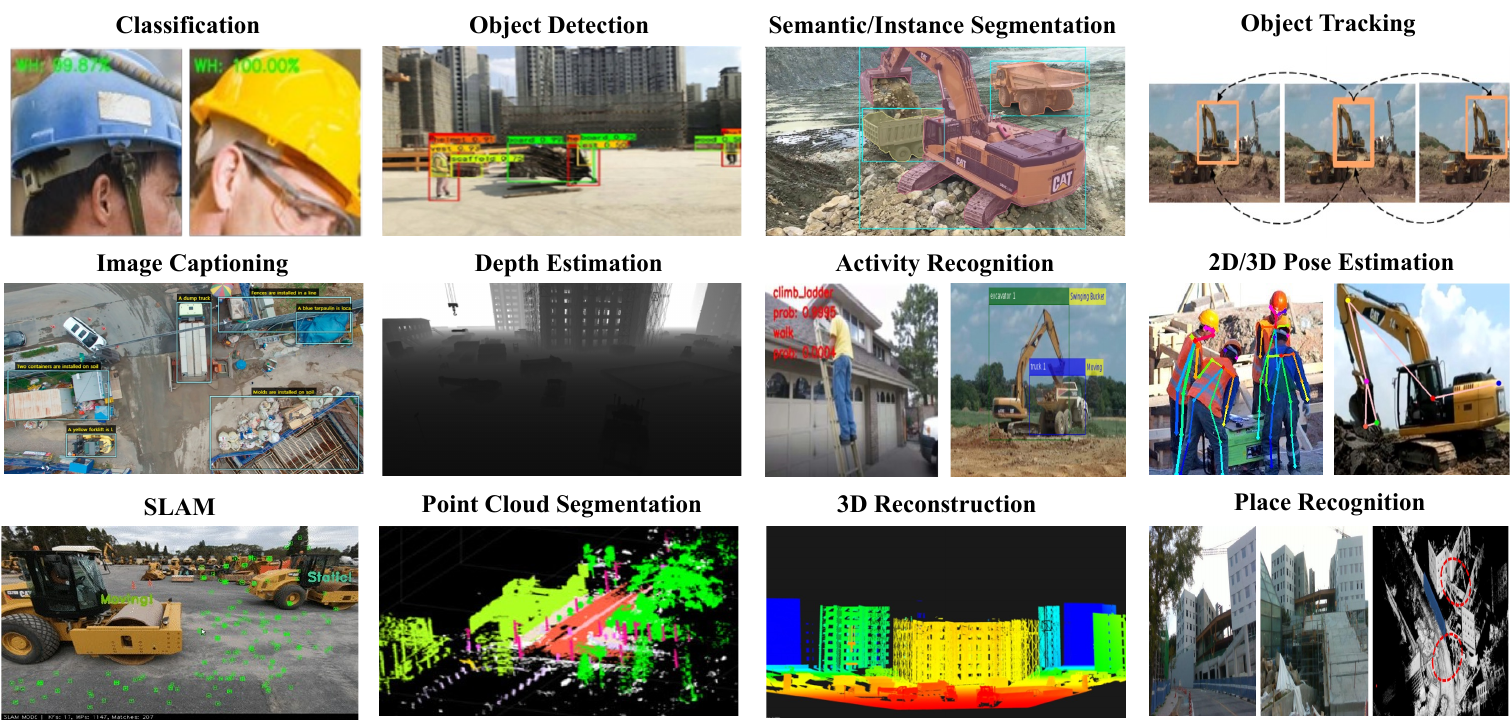} 
\caption{Examples of vision tasks in construction datasets. Sources (left to right, top to bottom): \textit{CPPE} \cite{xiong2021pose}; \textit{SODA} \cite{duan2022soda}; \textit{ACID} \cite{xiao2021development}; \textit{Earthmoving Equip. Tracking} \cite{roberts2019end}; \textit{UAV-acquired site image caption dataset} \cite{bang2020context}; \textit{VCVW-3D} \cite{ding2024virtual}; \textit{CMA} \cite{yang2023transformer} (left) and \textit{Earthmoving Equip. Tracking} \cite{roberts2019end} (right); \textit{CPPE} \cite{xiong2021pose} (left) and \textit{Equip. Pose Dataset} \cite{luo2020full} (right); \textit{kenki-posi Dataset} \cite{bao2021stereo}; \textit{PC-Urban} \cite{ibrahim2021annotation}; \textit{VCVW-3D} \cite{ding2024virtual}; \textit{ConPR} \cite{lee2024conpr}.}
\label{fig:task} 
\end{figure}

\paragraph{Vision Tasks} Fig.~\ref{fig:data_task_app} illustrates the mapping between datasets, task categories, and domain applications. Computer vision tasks represented in the datasets include classification, detection, segmentation, pose estimation, action recognition, image captioning, and 3D reconstruction (see Fig. \ref{fig:task}). Object detection is the most dominant vision task, appearing in 62\% of datasets (32/51). This prevalence reflects the central role of detection in identifying workers, equipment, and materials, forming the basis for safety compliance, utilization tracking, and progress monitoring. Semantic segmentation is included in 22\% (11/51), providing pixel-level details for tasks such as defect detection, material quality assessment, and site layout mapping. Image classification is supported by six datasets (12\%), including \textit{MINC-2500} \cite{bell2015material} for material recognition and \textit{BCS} \cite{cheng2024large} for building/site categorization.

Higher-level tasks are less common but add important capabilities. Pose estimation is supported by two datasets (4\%), such as \textit{Construction Equip. Pose} \cite{luo2020full} and \textit{CML} \cite{tian2022construction}, which support ergonomic evaluation and machinery orientation tracking. Action recognition is supported by datasets, including \textit{CMA} \cite{yang2023transformer} and \textit{Earthmoving Equip.} \cite{roberts2019end}, allowing temporal analysis of workflows and productivity. Image captioning, available in four datasets (8\%), links visual data with descriptive text, as in the \textit{Construction-Activity-Scenes Image Captioning Dataset} \cite{liu2020manifesting}, which supports automated documentation. 3D reconstruction and mapping tasks are represented by including \textit{VCVW-3D} \cite{ding2024virtual} and \textit{ConSLAM} \cite{trzeciak2023conslam}, contributing to site mapping and navigation.

\paragraph{Domain Applications in Datasets}
These vision tasks support several domain applications in construction monitoring (Fig.~\ref{fig:data_task_app}). The most widely supported is general site understanding, addressed by 18 datasets (35\%). They integrate multiple modalities and tasks, supporting scene-level reasoning and digital twin applications (e.g., \textit{VCVW-3D} \cite{ding2024virtual}). Safety monitoring follows closely, with 17 datasets (33\%) capturing PPE use, unsafe behaviors, and worker movements \cite{peng2018detecting,wang2020hardhat}. Progress tracking is enabled by 9 datasets (18\%), using object detection, captioning, and activity recognition to record task execution, equipment operations, and workflow dynamics (e.g., \textit{Earthmoving Equipment Tracking} \cite{roberts2019end}).

More specialized applications are less common. Quality control is addressed by 7 datasets (14\%), often using segmentation for defect detection or material classification (e.g., \textit{Concrete Crack Dataset} \cite{liu2019deep}). Equipment monitoring, supported by 4 datasets (8\%), leverages pose estimation and action recognition to evaluate machinery utilization and performance. Finally, site mapping and navigation, represented in 5 datasets (10\%), is supported by point clouds, mapping trajectories, and reconstruction datasets (e.g., \textit{PC-Urban} \cite{ibrahim2021annotation}).

\subsection{\textsc{OpenConstruction} Catalog Overview}
\label{sec:catalog}
The \textsc{OpenConstruction} catalog organizes open visual datasets reviewed into a central index to support transparent dataset discovery and reproducible benchmarking. Hosted on GitHub (\href{https://github.com/ruoxinx/OpenConstruction-Datasets}{https://github.com/ruoxinx/OpenConstruction-Datasets}), the catalog currently lists 51 datasets (2015–2024) collected from ground-level, UAV, and satellite platforms and spanning multiple modalities, including RGB imagery, thermal data, synthetic renderings, and 3D point clouds. 

Each dataset is described through a standardized JSON schema that records attributes essential for reuse and benchmarking: title and publication year; capture modality and platform; supported computer vision tasks (e.g., detection, segmentation, keypoints, captioning); construction-specific applications (e.g., safety monitoring, progress analysis, quality control); annotation types; dataset scale; typical resolution; license and access conditions; geographic provenance; and links to papers, code, and repositories. This schema enables systematic comparison, facilitates dataset selection for specific applications, and supports reproducible AI/ML experiments. To complement the online repository, Appendix~\ref{app:datasets} provides a summary table (Table~\ref{tab:datasets}) of the key attributes of all datasets.

To ensure responsible data governance, the catalog provides only bibliographic and descriptive metadata, such as titles, authors, DOIs, publication venues, and official repository links, without hosting or redistributing any underlying dataset or code. This approach preserves the intellectual property of dataset creators and ensures that all materials are accessed directly through the dataset owners’ official distribution channels.

The catalog is designed for community-driven sustainability, allowing practitioners and researchers to add or update entries through pull requests, with issue tracking to flag outdated links or inconsistencies. This collaborative approach promotes transparency, ensures traceability, and promotes the growth of a shared ecosystem of visual construction data.

\section{Discussion}
\label{sec:discuss}
This section discusses the gaps in current visual construction datasets and outlines a roadmap to strengthen open-access dataset development for robust AI/ML applications in construction monitoring.

\subsection{Gaps in Current Construction Open Visual Datasets}
Although recent developments in construction informatics have expanded the availability of open visual datasets, the current landscape does not yet provide the level of coverage, consistency, and governance required for AI/ML-ready data infrastructure in construction monitoring. Our review highlights four major categories of gaps that restrict the use of these datasets for reliable construction monitoring aimed at improving safety, efficiency, quality, and sustainability in diverse projects and site conditions (Table~\ref{tab:gaps}).

\begin{table}[htbp]
\centering
\caption{Major gaps in open-access visual datasets for construction monitoring}
\label{tab:gaps}
\renewcommand{\arraystretch}{1.2}
\begin{tabularx}{\textwidth}{p{2.8cm}X}
\toprule
\textbf{Category} & \textbf{Gap Implications in Construction Monitoring} \\
\midrule
\multirow{3}{=}{G1. Sensing \& Sampling} 
& G1.1 Limited multimodal and temporal coverage of dynamic activities \\ 
& G1.2 Regional and project bias reducing generalizability across jobsites\\ 
& G1.3 Privacy concerns of worker imagery \\
\midrule
\multirow{3}{=}{G2. Annotation \& Semantics} 
& G2.1 Incomplete metadata describing acquisition and site conditions \\ 
& G2.2 Lack of unified ontology for construction entities \\ 
& G2.3 Coarse annotations with limited segmentation or activity labels \\ 
\midrule
\multirow{3}{=}{G3. Context \& Interoperability} 
& G3.1 Missing contextual information (e.g., activity and weather)\\ 
& G3.2 Lack of standardized benchmarking protocols for domain tasks \\ 
& G3.3 Limited representation of environmental challenges such as occlusion, lighting, and weather variability \\ 
\midrule
\multirow{3}{=}{G4. Governance, Access \& Reuse} 
& G4.1 Ambiguous or restrictive dataset licensing for research and practices \\ 
& G4.2 Limited dataset discoverability and long-term maintenance \\ 
& G4.3 Lack of construction-specific data-sharing policies and stakeholder participation in governance\\ 
\bottomrule
\end{tabularx}
\end{table}

\subsubsection{Gap 1: Limited Sensing and Sampling of Dynamic Jobsite Environments}
The majority of datasets focus on a narrow range of modalities and lack temporal depth. Of the 51 corpora reviewed, 46 (90.2\%) rely primarily on RGB imagery, while only a small fraction includes thermal (3.9\%), RGB-D (3.9\%), or 3D point cloud data (5.9\%). Temporal information is also limited: only 7 datasets (13.7\%) contain videos or activity trajectories, restricting the ability to model evolving workflows, worker--equipment interactions, and near-miss events. Geographic diversity is constrained, as most datasets originate from single projects or localized regions, which could restrict the generalizability of trained models to diverse construction practices and environments. Privacy considerations are inconsistently addressed, as identifiable worker imagery is rarely anonymized or blurred, creating potential ethical and legal barriers to reuse.

\subsubsection{Gap 2: Limited Annotation and Semantics for Construction Activities}
Annotation types among the reviewed datasets show a high concentration in bounding-box labeling, which appears in 40 of the 51 datasets (78\%). Pixel-level segmentation is available in 11 datasets (22\%), supporting detailed analyses such as defect detection or material assessment. Documentation of annotation processes is also inconsistent. While some datasets report labeling guidelines or validation procedures (e.g., CML \cite{tian2022construction}), many provide only limited detail on annotation protocols or quality checks. As a result, information on inter-annotator agreement, expert validation, or error analysis is rarely available in the dataset descriptions. Metadata descriptors are often sparse, and acquisition conditions (e.g., lighting, weather, project phase) rarely are specified. Terminology varies across datasets, such as “helmet” versus “hardhat” or generic versus detailed equipment categories, complicating dataset comparability.

\subsubsection{Gap 3: Limited Process Context and Interoperability with Construction Workflows}
Contextual metadata in current visual datasets is often sparse. Key descriptors such as weather, lighting, or site activity conditions are rarely included in the dataset documentation, limiting the ability to align visual data with construction planning, scheduling, or digital twin systems. Benchmarking practices are also fragmented. Multiple datasets target similar tasks, such as PPE detection \cite{peng2018detecting,wang2020hardhat} or site segmentation \cite{wang2022deep,xuehui2021dataset}, yet no unified or widely adopted protocols exist for standardized evaluation. Furthermore, environmental complexities common to active jobsites, such as occlusions, variable illumination, and dynamic site conditions, are rarely represented. Only a few datasets document acquisition under various weather or lighting conditions. These limitations hinder robust model assessment and reduce the reliability of AI/ML deployment across heterogeneous construction environments.

\subsubsection{Gap 4: Limited Governance, Access, and Reuse for Practical Adoption}
Licensing and governance practices across the datasets vary considerably. Of the 51 reviewed corpora, 13 (25\%) do not specify a license. The absence of explicit licensing in a substantial portion of datasets creates uncertainty about reuse, particularly in applied or cross-institutional projects. Discoverability also remains a challenge. The construction industry lacks centralized repositories comparable to those established in fields such as medical imaging or autonomous driving \cite{diaz2021data,liu2024survey}. As a result, indexing is scattered across individual publications, platforms, and institutional repositories, with limited long-term maintenance guarantees. In addition, few datasets explicitly describe governance structures or community participation in their curation. These factors constrain dataset reuse and slow the establishment of standardized benchmarks for construction monitoring.

\subsection{Roadmap: Fostering Open-Access Data Infrastructure for Construction Monitoring}
The analysis of current open visual datasets highlights critical limitations that restrict their broader impact on research and practice. The FAIR principles, emphasizing findability, accessibility, interoperability, and reusability, provide a widely adopted foundation for scientific data stewardship \cite{wilkinson2016fair}. FAIR principles emphasize that datasets should be discoverable through standardized metadata, openly accessible under transparent conditions, interoperable across systems through consistent formats, and reusable with clear documentation and licensing. 

\begin{table}[htbp]
\centering
\caption{Summary of frameworks and principles for data collection, sharing, and management}
\label{tab:frameworks}
\resizebox{\textwidth}{!}{%
\begin{tabular}{p{7cm} p{5cm} p{6.5cm}}
\toprule
\textbf{Framework} & \textbf{Principles} & \textbf{Application in Practice} \\
\midrule
FAIR Data Principles \cite{wilkinson2016fair} & Findability, Accessibility, Interoperability, Reusability & Metadata-rich documentation and open formats improve discoverability, integration, and reuse. \\

ISO 8000: Data Quality and Master Data \cite{iso8000} & Accuracy, Consistency, Completeness, Quality Assurance & Provides a framework for defining, measuring, and assuring data quality. \\

ISO/IEC 27001: Information Security Management \cite{iso27001} & Protection, Controlled Access & Sets standards for secure storage, encryption, and access control of sensitive project data. \\

Construction-Operations Building Information Exchange (COBie) \cite{cobie} & Structured Data Delivery, Asset Information Exchange & Specifies standardized formats for delivering equipment lists, warranties, and maintenance data at project handover to facility managers. \\

ISO 19650: Organization and Digitization of Information using BIM \cite{iso19650} & Standardization, Lifecycle Integration, Transparency & Consistent naming, metadata, and workflows; managed through a Common Data Environment (CDE). \\

CIC BIM Protocol \& AIA Document E203-2013 \cite{cicBIMprotocol,aiaE203} & Clarity, Licensing, Rights Protection & Define ownership and licensing of project data; set rules for sharing and liability in BIM collaboration. \\
\bottomrule
\end{tabular}%
}
\end{table}

\begin{figure}[ht]
    \centering
    \includegraphics[width=0.98\linewidth]{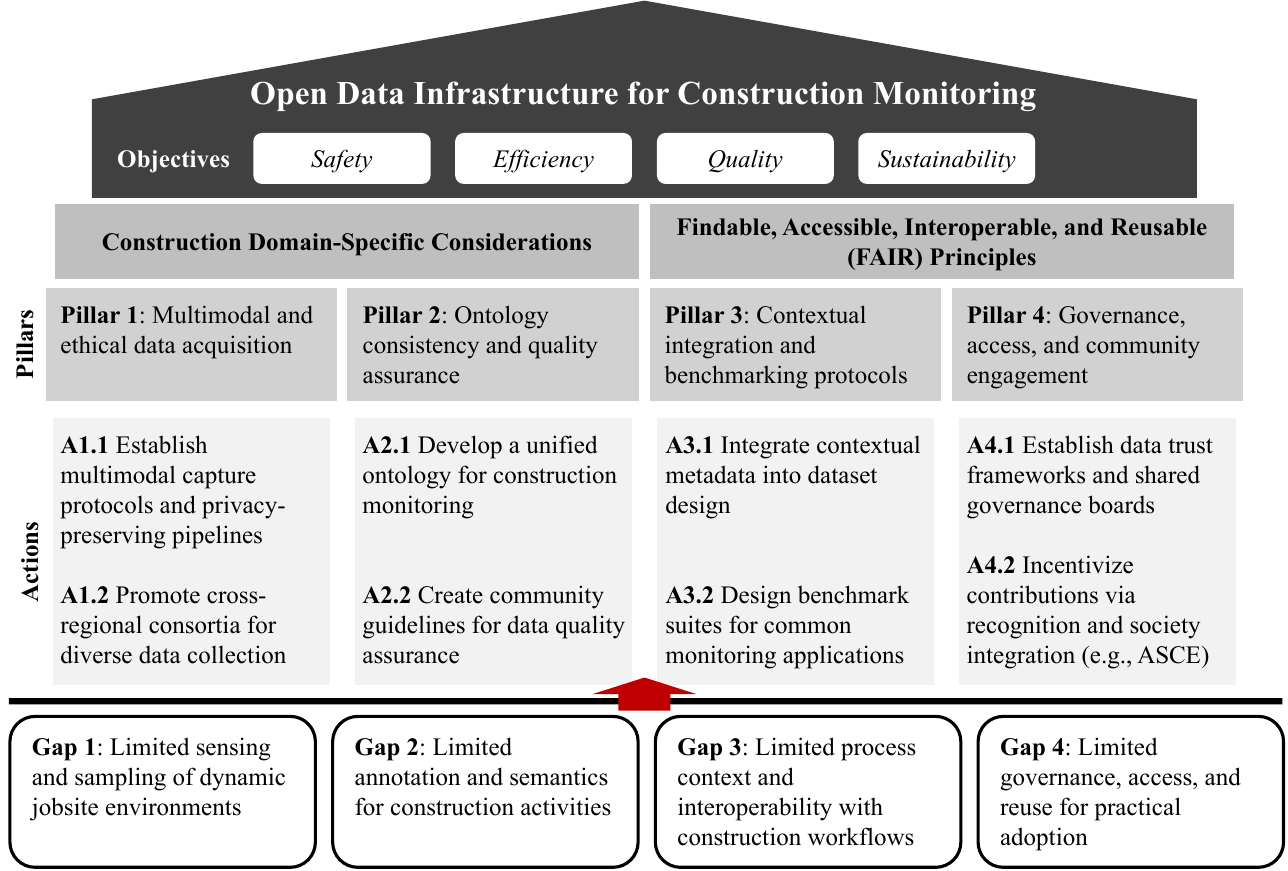}
    \caption{Proposed roadmap for advancing open-access data infrastructure in construction monitoring}
    \label{fig:open_framework}
\end{figure}

Although these principles establish an important baseline, previous studies and frameworks in construction informatics have pointed out that sector-specific challenges extend beyond general data management. Issues such as protecting worker privacy, maintaining annotation consistency, embedding contextual information (e.g., environment and site conditions), and establishing governance mechanisms are critical to ensure that datasets support safe, efficient, and sustainable practices in construction monitoring (Table~\ref{tab:frameworks}). Building on this foundation, our roadmap (Fig.~\ref{fig:open_framework}) integrates FAIR with construction-specific considerations to strengthen open-access data infrastructure.

\subsubsection{Pillar 1: Multimodal and Ethical Data Acquisition}
Advancing construction monitoring requires datasets that reflect the multimodal and dynamic nature of jobsites. Future initiatives should establish acquisition protocols that integrate various data streams in complementary ways \cite{bayoudh2022survey}. Professional societies such as the American Society of Civil Engineers (ASCE) could coordinate the development of capture guidelines that ensure comparability across projects and platforms. Research institutions and industry partners may contribute through joint pilot studies that collect multimodal data under varying conditions.

Ethical safeguards should also be prioritized. Privacy-preserving pipelines, such as face blurring, skeletal representations of workers, or consent protocols, should be incorporated into data curation pipelines \cite{zhao2025visual}. Institutional Review Boards (IRBs), together with construction firms, can provide oversight to ensure ethical alignment. Incorporating these measures will expand the representativeness of open-access datasets while fostering trust and accountability in their use.

\subsubsection{Pillar 2: Ontology Consistency and Quality Assurance}
Interoperability in construction monitoring datasets should be advanced through a unified ontology that adapts industry classification systems (e.g., IFC and OmniClass) to represent jobsite entities such as workers, equipment, temporary structures, and activity states \cite{xiong2022facilitating}. This shared framework enables consistent data representation and integration across platforms. Beyond ontology alignment, annotation quality can be supported through structured guidelines and validation processes. This effort could be advanced by professional societies (e.g., ASCE), publication communities that set data standards, and industry consortia that provide domain expertise. Potential mechanisms include open annotation handbooks, inter-rater reliability protocols, and organizing benchmarking challenges for systematic validation \cite{ehrlinger2022survey}.

The success of these initiatives can be measured using established metrics: syntactic conformance (compliance with schema), semantic alignment (mapping of labels to shared vocabularies), and annotation reliability \cite{zhu2014assessing}. Embedding such criteria into dataset practices will promote consistency, reproducibility, and trust in open visual resources for construction monitoring.

\subsubsection{Pillar 3: Contextual Integration and Benchmarking Protocols}
The long-term usefulness of visual datasets is enhanced when they are integrated into construction workflows and evaluated under realistic conditions \cite{elkhidir2025toward}. This involves systematic inclusion of contextual metadata, such as project phase, site type, and environmental conditions, so that datasets can be meaningfully linked with BIM models, schedules, and digital twins. Such integration can be advanced by adapting existing construction vocabularies to practical metadata templates, piloting their use on collaborative projects, and publishing implementation guidelines to ensure compatibility with established construction data exchange standards.

Benchmarking can also be strengthened through coordinated evaluation practices. ASCE technical committees and industry partnerships may define common task benchmarks (e.g., equipment tracking and compliance monitoring) and curate test suites that reflect jobsite challenges such as occlusion, variable lighting, and weather effects. Evaluation methods could combine leaderboards to track algorithm performance with shared baselines that promote transparency and comparability \cite{beyer2019reliable}. These efforts would provide comparable performance metrics and encourage the development of models that are robust and broadly applicable between projects and regions.

\subsubsection{Pillar 4: Governance, Access, and Community Engagement}
The long-term sustainability of open-access datasets depends on governance structures that move beyond isolated project-level releases. Governance boards established and coordinated by professional societies such as ASCE could play a central role by issuing standardized licensing templates to reduce ambiguity, maintaining repositories with long-term hosting commitments, and providing procedures for resolving intellectual property or licensing disputes \cite{al2018exploring}.

Community engagement mechanisms are essential to ensure continuity and growth. Professional societies and consortia can foster participation by establishing formal recognition mechanisms for contributors, convening workshops to advance dataset practices, and fostering partnerships between academic researchers and industry stakeholders \cite{sansone2019fairsharing}. Embedding such governance and engagement practices into the research and practice ecosystem will help make construction monitoring datasets sustainable, trustworthy, and aligned with the FAIR principles.

\section{Conclusion}
\label{sec:conclude}
This review synthesized 51 open visual datasets for construction monitoring (2015–2024), capturing temporal and geographic trends, sensing modalities, acquisition methods, annotations, tasks, and domain applications. The analysis shows that most resources remain RGB-centric and focused on object detection, while temporal data (video), multimodal sensing (e.g., thermal and point clouds), and higher-level tasks (e.g., action and captioning) are comparatively scarce. Annotation practices also vary in depth and transparency, with limited reports of inter-rater agreement or expert validation. Licensing is heterogeneous and often unspecified, creating uncertainty for reuse. These constraints could restrict the generalizability and long-term utility of existing datasets.

The analysis highlights four priority areas for advancing construction monitoring: broader sensing coverage, more consistent and detailed annotations, richer contextual and benchmarking information, and stronger provisions for governance and accessibility. To support these developments, we proposed a roadmap aligned with the FAIR principles and domain-specific considerations. First, multimodal and ethical acquisition should expand beyond RGB to integrate thermal, depth, point cloud, and video, supported by privacy safeguards and ethical oversight. Second, ontology consistency and quality assurance should be advanced through unified vocabularies and validated annotation protocols. Third, contextual integration and benchmarking can be advanced by adding basic metadata such as project phase or environmental conditions, and by gradually developing test suites that account for common challenges such as occlusion, lighting, and weather. Finally, governance, access, and community engagement should be strengthened through standardized licensing, sustainable repositories, and recognition programs that maintain contributions over time.

These directions also have practical implications for industry stakeholders. Contractors and owners can leverage richer datasets for more reliable safety monitoring, productivity analysis, and progress verification. Standards bodies and technology providers benefit from improved interoperability, reducing integration costs and enabling broader deployment of digital tools. For professional societies, stronger governance and benchmarking frameworks provide opportunities to set standards and foster trust across the sector. By working together, researchers and practitioners can strengthen the foundations of open-data infrastructure in construction monitoring and create resources that advance safety, efficiency, and resilience in the built environment.

\noindent{\textbf{Acknowledgments}}
This research was supported by startup funding from Bert S. Turner Department of Construction Management at Louisiana State University (LSU) and the Cajun Industries Professorship in Construction Management. The findings, interpretations, and conclusions expressed in this study do not necessarily reflect the views of Cajun Industries or LSU.







%
%
\bibliography{ascexmpl-new}
%
\appendix
\section{The \textsc{OpenConstruction} Catalog: Open-access Visual Datasets for Construction Monitoring}
\label{app:datasets}

\noindent This appendix provides an overview of the \textsc{OpenConstruction} Catalog, a curated collection of visual datasets that support research on construction monitoring. Table~\ref{tab:datasets} summarizes key characteristics of each dataset, including data type, collection method, dataset size, annotation format, supported tasks, license, and access information.

\begin{landscape}   
\scriptsize
\begin{longtable}{p{3cm}p{0.5cm}p{1.8cm}p{2.2cm}p{1.6cm}p{1cm}p{1.5cm}p{2.5cm}p{2.7cm}p{1.2cm}p{0.8cm}}
\caption{A summary of publicly available visual datasets for construction monitoring. Note: N/A: Information not available or not specified.}
\label{tab:datasets} \\
\toprule
Dataset Name & Year & Data Type & Collection Method & Images (n) & Classes & Resolution & Annotation Type & Construction Element & License & Access \\
\midrule
\endfirsthead
\multicolumn{11}{c}{Table \ref{tab:datasets} A summary of publicly available visual datasets for construction monitoring and analysis (continued)} \\
\toprule
Dataset Name & Year & Data Type & Collection Method & Images/Videos (n) & Classes & Resolution & Annotation Type & Construction Element & License & Access \\
\midrule
\endhead
\midrule
\multicolumn{11}{r}{Continued on next page} \\
\endfoot
\bottomrule
\endlastfoot
MINC-2500 \cite{bell2015material} & 2015 & RGB & Web-scraped & 57,500 & 23 & 362×362 & Image classification & Construction materials & CC BY 4.0 & Public \\

SCUT-HEAD \cite{peng2018detecting} & 2018 & RGB & Web-scraped & 4,405 & 1 & Variable & 2D bounding box & Worker detection & N/A & Public \\

GDUT-HWD \cite{wu2019automatic} & 2019 & RGB & Web-scraped & 3,174 & 5 & Variable & 2D bounding box & Safety equipment & Apache-2.0 & Public \\

Concrete Crack \cite{liu2019deep} & 2019 & RGB  & Hybrid (web+field) & 11,300 & 2 & 256×256 & Pixel segmentation & Structural defects & CC BY 4.0 & Public \\

Earthmoving Equipment Tracking \cite{roberts2019end} & 2019 & RGB & Fixed ground camera & 515 (video clips) & 7 & 480×272 & 2D bounding box and action classification & Equipment activities & CC BY 4.0 & Public \\

SHWD \cite{njvision2019} & 2019 & RGB & Web-scraped & 7,581 & 2 & Variable & 2D bounding box & Safety equipment & MIT & Public \\

ACID \cite{xiao2021development,xiao2022deep} & 2020 & RGB & Multi-platform (UAV+ground) & 10,000 & 10 & 608×608 & 2D bounding box, pixel segmentation and image caption & Construction equipment & CC BY-NC 4.0 & Request \\

Construction equipment Pose \cite{luo2020full} & 2020 & RGB & Web-scraped & 1,281 & 6 & 512×512 & 2D machine keypoint & Equipment pose (excavator) & CC BY 4.0 & Public \\

Hardhat-Wearing \cite{wang2020hardhat} & 2020 & RGB & N/A & 7,064 & 2 & Variable & 2D bounding box & Safety equipment & CC0 1.0 & Public \\

UAV-acquired site image caption dataset \cite{bang2020context} & 2020 & RGB &UAV (aerial) & 1,431 & N/A & 960×540 & Image caption & Site activities & CC BY 4.0 & Public \\

Construction-activity-scenes-image-captioning dataset \cite{liu2020manifesting} & 2020 & RGB & Fixed ground camera & 7,382 & N/A & Variable & Image caption & Construction activities & N/A & Public \\

MOCS \cite{xuehui2021dataset} & 2021 & RGB & Multi-platform (UAV+ground) & 41,668 & 13 & Variable & 2D bounding box and pixel segmentation & Site objects & CC BY-NC 4.0 & Request \\

Peru Construction \cite{del2022dataset} & 2021 & RGB & Site camera & 1,046 & 8 & Variable & 2D bounding box & Equipment and workers & CC BY 4.0 & Public \\

PC-Urban \cite{ibrahim2021annotation} & 2021 & 3D point clouds & LiDAR sensor & N/A & 25 & N/A & 3D segmentation & Urban elements & CC BY 4.0 & Public \\

kenki-posi \cite{bao2021stereo} & 2021 & RGB & Stereo camera & N/A & N/A & 960×540 & Camera pose & Machine positioning & GNU GPL 3.0 & Public \\

CHV \cite{wang2021fast} & 2021 & RGB & Curated compilation & 1,330 & 6 & Variable & 2D bounding box & Safety equipment & N/A & Public \\

CDE \cite{xiong2021machine} & 2021 & RGB + synthetic & Hybrid (real+synthetic) & 4,875 & 1 & Variable & 2D bounding box & Environmental monitoring & MIT & Public \\

CPPE \cite{xiong2021pose} & 2021 & RGB & Web-scraped & 932 & 3 & Variable & 2D bounding box & Safety equipment & MIT & Public \\

Safety harness \cite{xu2023novel} & 2022 & RGB & Hybrid (web+field) & 3,300 & 3 & Variable & 2D bounding box & Safety equipment & N/A & Public \\

CML \cite{tian2022construction} & 2022 & RGB + RGB-D & Hybrid+IMU & 146,480 & 225 & Variable & 3D keypoints & Worker activities & CC BY 4.0 & Public \\

Opening Seg. \cite{pantoja2022generating} & 2022 & RGB & Camera & 270 & 3 & Variable & Pixel segmentation & Building elements & CC BY 4.0 & Public \\

SODA \cite{duan2022soda} & 2022 & RGB & Multi-platform (UAV+ground) & 19,846 & 15 & $\geq$ 1920×1080 & 2D bounding box & Site objects & N/A & Public \\

CHVG \cite{ferdous2022ppe} & 2022 & RGB & Hybrid (web+field) & 1,699 & 8 & 640×640 & 2D bounding box & Safety equipment & CC BY 4.0 & Public \\

SHEL5K \cite{otgonbold2022shel5k} & 2022 & RGB & Curated compilation & 5,000 & 6 & Variable & 2D bounding box & Safety equipment & CC BY 4.0 & Public \\

Site Segmentation \cite{wang2022deep} & 2022 & RGB & Web-scraped & 859 & 12 & Variable & Pixel segmentation & Site elements & MIT & Public \\

BCS \cite{cheng2024large} & 2023 & RGB & Hybrid (web+field) & 212,000 & 61 & Variable & Image classification & Building and site types & N/A & Request \\

SMART Dataset \cite{goldberg2023automated} & 2023 & RGB & Satellite & 13,000 & 5 & Variable & 2D bounding box & Construction stages & MIT & Public \\

CODD \cite{demetriou2024codd} & 2023 & RGB & Site camera & 3,129 & 10 & 1920×1200 & 2D bounding box and pixel segmentation & Construction waste & CC BY 4.0 & Public \\

ConSLAM \cite{trzeciak2023conslam} & 2023 & RGB + 3D point clouds & Multi-sensor & N/A & N/A & Variable & Trajectory & Site mapping & N/A & Public \\

CIS \cite{yan2023construction} & 2023 & RGB & Multi-platform & 50,000 & 10 & Variable & Pixel segmentation & Site objects & CC BY-NC 4.0 & Request \\

Visible-Dust \cite{wang2023deep} & 2023 & RGB & Camera & 7,500 & 9 & Variable & 2D bounding box & Dust and prevention & MIT & Public \\

CWT \cite{guan2021tns} & 2023 & RGB & Multi-sensor & N/A & N/A & Variable & Terrain mapping & Site conditions & N/A & Request \\

Construction material textual dataset \cite{mengiste2024transfer} & 2023 & RGB & Site camera & 208 & 7 & Variable & Image classification & Material conditions & CC BY 4.0 & Public \\

CMA \cite{yang2023transformer} & 2023 & RGB & Web-scraped & 1,595 (video clips) & 7 & Variable & Action classification & Worker behaviors & MIT & Public \\

VCVW-3D \cite{ding2024virtual} & 2023 & Synthetic + RGB-D & Unity engine & 375,000 & 10 & 1920×1080 & 2D/3D bounding box, pixel segmentation, and depth maps & Site objects & N/A & Public \\

Scaffolding \cite{Chowdhury2023video} & 2023 & RGB & Hybrid (web+field) & 3,040 & 15 & Variable & 2D bounding box & Scaffolding detection & CC0 1.0 & Public \\

Small tool image dataset \cite{lee2023small} & 2023 & RGB & Site camera & 25,084 & 12 & Variable & 2D bounding box & Construction tools & CC BY 4.0 & Public\\

ConPR \cite{lee2024conpr} & 2023 & RGB + 3D point clouds & Multi-sensor & N/A & N/A & Variable & Place recognition & Site recognition & CC BY-NC 4.0 & Public \\

Thermal Safety \cite{yirong2023thermal} & 2023 & Thermal & Thermal camera & 2,000 & 4 & 200×200 & Image classification & Worker actions & CC BY 4.0 & Public \\

ThermalNet Dataset \cite{wu2023thermal} & 2023 & Thermal & Thermal camera & 19,418 & 7 & 480×640 & Image classification & Worker hand gesture recognition & N/A & Public \\

Synthetic Rebar \cite{wang2023synthetic} & 2023 & Synthetic & Unity engine & 2,500 & 1 & Variable & Pixel segmentation & Rebar detection & N/A & Public \\

Rebar Counting \cite{wang2024labelled} & 2024 & RGB & UAV & 728 & 1 & 1500×900 & 2D bounding box & Rebar inspection & CC BY 4.0 & Public \\

ROADWork Dataset \cite{ghosh2024roadwork} & 2024 & RGB & Hybrid (web+field) & 7,416 & N/A & Variable & Object annotations, scene tags, text descriptions, 2D/3D drivable paths & Work zones & ODC-BY & Public \\

AIDCON \cite{ersoz2024aidcon} & 2024 & RGB & UAV & 2,155 & 9 & Variable & 2D bounding box and pixel segmentation & Construction machines & CC BY-NC 4.0 & Request \\

Steel Cross-Section Dataset \cite{chen2024automated} & 2024 & RGB & Smartphone & 3,113 & 4 & Variable & 2D bounding box & Steel components & N/A & Public \\

Building Char. \cite{wang2024building} & 2024 & RGB & Street view & 3,717 & 15 & 640×640 & Image classification & Building features & CC BY 4.0 & Public \\

ConRebSeg \cite{schmidt2024conrebseg} & 2024 & RGB & Hybrid (web+field) & 14,805 & 4 & 1920×1200 & 2D bounding box and pixel segmentation & Reinforcement elements & CC BY 4.0 & Public \\

VL-Con Dataset \cite{hsu2024vl} & 2024 & RGB & Curated compilation & 4,073 & 641 & Variable & Image-text pairs & Site monitoring & Apache-2.0 & Public \\

Object Detection at Construction Sites \cite{jongho2024development} & 2024 & RGB & Hybrid (web+field) & 87,766 & 12 & Variable & 2D bounding box & Site objects & CC BY 4.0 & Public \\

SAFE \cite{zhai2023extracting} & 2024 & RGB & Hybrid (web+field) & 5,659 & N/A & 256×256 & Image caption & Safety behaviors & N/A & Request \\

Small-sized tool dataset \cite{han2024utilizing} & 2024 & RGB + synthetic & Hybrid (real+synthetic) & 8,576 & 4 & Variable & 2D bounding box & Construction tools & CC BY-NC 4.0 & Request \\
\end{longtable}
\end{landscape}
\end{document}